# Structural Damage Detection

# Using Randomized Trained Neural Networks

**Ismoyo Haryanto\***, **Joga Dharma Setiawan\*,** and **Agus Budiyono** ⁼

\*Mechanical Engineering Department  
Diponegoro University, Semarang, Indonesia.  
e-mail: ismoyo@mesin.ft.undip.ac.id

⁼Aeronautics and Astronautics Department  
Bandung Institute of Technology, Bandung, Indonesia.  
e-mail: agus.budiyono@ae.itb.ac.id

**Abstract**

A computationally method on damage detection problems in structures was conducted using neural networks. The problem that is considered in this works consists of estimating the existence, location and extent of stiffness reduction in structure which is indicated by the changes of the structural static parameters such as deflection and strain. The neural network was trained to recognize the behaviour of static parameter of the undamaged structure as well as of the structure with various possible damage extent and location which were modelled as random states. The proposed techniques were applied to detect damage in a simply supported beam. The structure was analyzed using finite-element-method (FEM) and the damage identification was conducted by a back-propagation neural network using the change of the structural strain and displacement. The results showed that using proposed method the strain is more efficient for identification of damage than the displacement.

## 1 Introduction

Structural systems or machinery components tend to get damage during their operation life. Therefore, an effective and reliable damage assessment methodology of the structural system is valuable tool. A determination of safety level of a structural system during its operational life is essential not only for safe operation but also maintenance cost reduction and failure prevention.

Occurring of damage in a structural element reduces stiffness of the structure and generates a small perturbation in its static or dynamic responses. A perturbation on static responses can be identified by the behavior of either displacements or strains. Meanwhile, the behavior of natural frequencies and mode shapes can be used to identify the perturbation on dynamic responses of the structure. A combination of measured response and finite-element-methods (FEM) then can be developed in order to identify these response perturbations which can be used to determine the size and location of the damage of the structure.

Response of damaged structure will follow the pattern of the size and location of the damage on its structure. Bishop has shown that this pattern can be generalized using Artificial Neural Network/NN [1]. Therefore, the damage detection on a structural system or a machinery component can be conducted using NN which was trained to identify the pattern of response characteristic of the structure.

Maity and Saha have developed a damage assessment in structure from changes in static parameter using approach [8]. Unfortunately, this assessment was focused on single element damage and multiple element damage which consists only of two damaged elements. In practical point of view this methodology is inadequate. Therefore, a more general damage assessment methodology has to be developed.

The objective of this research is to develop a structural damaged detection methodology from changes in static parameter, i.e.: displacement and strain of a simple cantilever beam using neural network combined with FEM. In present work a random state is proposed to simulate stiffness reduction factor and damage location of the structure such that values of the stiffness reduction factor and the damage location are random. Using this random state the proposed method of structural damage assessment may be able to be applied in more general condition.

## 2 Problem Formulation

First step in damage detection of a structure using neural network is modeling of the structure to obtain data set which is used as input in the network training. This structural modeling has to be able to represent all possibilities of damage condition on the structure. The damage of the structure is modeled by stiffness reduction and consists of size and location on the structure. In order to obtain the data set as input for network training, values of the stiffness reduction are assumed to be random number between 0 and 1. The number and the location of damaged structural element are also assumed to be random and it may be multiple element damage. In this present work, structural response of strain and displacement due to specific loading obtained by FEM were chosen as data set used for training the network.

When the structural responses as input data set was obtained then training of the network is conducted until outputs of the networks satisfy the desired target or until the network reach desired performance which is indicated by error level (difference between output and desired target of network). Usually this error is formulated as mean square error (MSE). The above principle of neural network is illustrated by a simple schematic in Fig. 1 below.

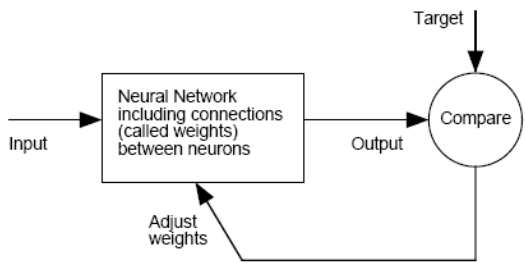

**Fig. 1. A schematic of an artificial neural network [2]**

At following section the structural modelling and the neural networks will be discussed more detail.

**2.1 Structural Modeling**
The strain energy of a structural element is formulated as

$$U = \frac{1}{2}\int_{Vol} \{\sigma\}^T \{\varepsilon\} dVol \qquad (1)$$

Introducing the stress-strain relation in Eq. (1), we have

$$U = \frac{1}{2}\int_{Vol} \{\varepsilon\}^T [D]\{\varepsilon\} dVol \qquad (2)$$

Note that $\{\sigma\} = [D]\{\varepsilon\}$, whereas $D$ represents constitutive matrix of the structure material. On the other hand the strain-displacement relation is given by

$$\{\varepsilon\} = [B]\{d\} \qquad (3)$$

where $B$ is derivative of shape function and $d$ is structural displacement. Finally the stiffness matrices of the structural element are formulated as

$$[K] = \int_{Vol} [B]^T [D][B] dVol \qquad (4)$$

and the strain of the structure is given by the following formulae [6]

$$\varepsilon(x,z) = \frac{du}{dx} = -z\frac{d^2v}{dx^2} = -zv'' \qquad (5)$$

where $u$ and $v$ are vertical and transversal displacement, respectively. Whereas $x$ and $z$ are structural coordinate in horizontal and vertical direction as well as.

The nodal displacement due to applied load then can be calculated using following relation

$$[K]\{d\} = \{F\} \qquad (6)$$

where $\{F\}$ is the applied load node.

The damaged modelling on the structure can be conducted by a reduction of the structural stiffness which is represented by reduction of the cross section area of the structural elements. In structural modelling using FEM the stiffness matrices of the damaged structural element are formulated as follows:

$$[K_d] = ee[K] \qquad (7)$$

where:
$[K_d]$ : stiffness matrix of damaged structural element
$[K]$ : stiffness matrix of undamaged structural element
$ee$ : stiffness reduction factor

It should be noted that the value of stiffness reduction factor is between zero and one ($0 < ee \leq 1$). For undamaged structural element $ee = 1$, meanwhile $ee < 1$ represents damaged structural element.

**2.2 Neural Network**
An Artificial Neural Networks (ANN) is computational system which is inspired by the biological brain in their structure, data processing and learning ability with some assumptions as follows [2][5][8][9]:

- The information processing are conducted at the simple element called neuron
- The signals are transmitted from neuron to other neuron through the connection
- Each connection between neurons has a specific weighting factor
- In order to determine the output, an activation function is applied in the input and then the output of the system compared to a desired target.

Generally, the neural network is characterized by: network architecture which simulates relation pattern of neurons, activation function, and training method. Fig.2 illustrates multilayer network, the most common architecture of network. Besides input and output this network also consists of hidden layers although it is possible to build a network without hidden layer. The network in Fig. 2 consists of $R$ number of input unit ($p_1$, $p_2$, ... , $p_R$), two hidden layers and one output layer.

The input layer receives input pattern and transmits the signal directly to the next layer. Meanwhile, the hidden layer consists of a certain number of processing units and each node in the preceding layer is fully connected to all processing units. These connections are called the weights that represent different weighting scales to the input signals. The weighted signals then are summed up by the processing unit and a response transmitting is activated to the next layer. The activation function may be linear or non-linear function. The above procedure is illustrated in Fig 3.

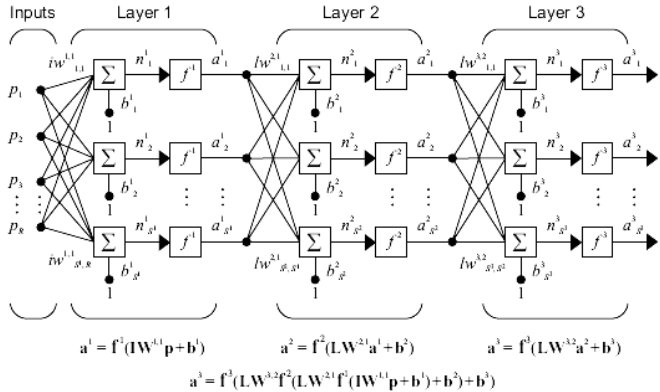

Fig 2. Multilayer network [2]

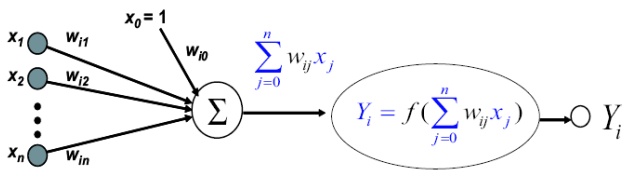

Fig 3. Artificial neuron

The input pattern is propagated forward and actual responses are obtained. The difference between the actual and the desired outputs is then through network propagated backward to modify the weights such that the mean-square error (MSE) is minimized. This supervised training continues until the training process complete. In present work, Lavenberg Merquardt algorithm was applied as training algorithm.

Using Lavenberg Merquardt algorithm the mean square error (MSE) is used as performance function, which is formulated as follows

$$MSE = \frac{1}{N}\sum_{i=1}^{N}(e_i)^2 = \frac{1}{N}\sum_{i=1}^{N}(y_i - t_i)^2 \quad (8)$$

where $t_i$ is desired target, $y_i$ is actual output, and $N$ is number of training data. According to Lavenberg-Marquardt [2][4]

$$\Delta w = [J^T(w).J(w)+\mu I]^{-1} J^T e(w) \quad (9)$$

where $J(w)$ is Jacobian matrix

$$J(w) = \begin{bmatrix} \frac{\partial e_1(w)}{\partial w_1} & \frac{\partial e_1(w)}{\partial w_2} & \cdots & \frac{\partial e_1(w)}{\partial w_n} \\ \frac{\partial e_2(w)}{\partial w_1} & \frac{\partial e_2(w)}{\partial w_2} & \cdots & \frac{\partial e_2(w)}{\partial w_n} \\ \vdots & \vdots & \ddots & \vdots \\ \frac{\partial e_N(w)}{\partial w_1} & \frac{\partial e_N(w)}{\partial w_2} & \cdots & \frac{\partial e_N(w)}{\partial w_n} \end{bmatrix} \quad (10)$$

The Marquardt modification to the back-propagation algorithm thus proceeds as follows [4]:

1. Present all inputs to the network and compute the corresponding network outputs and MSE over all inputs (using Eq. (8)).
2. Compute the Jacobian matrix Eq. (10).
3. Solve (9) to obtain $\Delta w$
4. Recompute MSE using $w+\Delta w$. If this new MSE is smaller than that computed in step 1, then reduce μ by β, and go back to step 1. If the MSE is not reduced, then increase μ by β and go back to step 3.
5. The algorithm is assumed to have converged when the MSE has been reduced to some error goal.

## 3  Numerical Example

The computer codes which have been developed in present work were applied on a simple cantilever beam structure. For training the neural network the calculated static displacement and strain at several nodal points were used. Reducing of *EI* values is applied in order to define the damage of the structural element.

Fig. 4 shows a cantilever beam with rectangular cross-section subjected to vertical load of 100 N at the tip. The beam has 0.2 m in length, uniform cross section with $h$ = 0.01 m and $b$ = 0.02 m, whereas the Young's Moduli of the material is $E$ = 200 GPa. The beam was divided into eight elements to find the deflection and the strain using FEM.

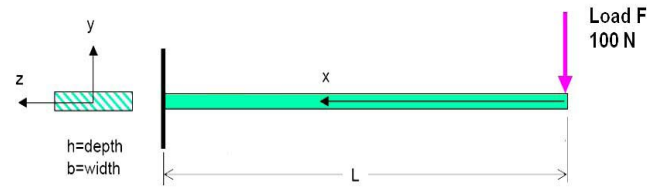

Fig. 4 Cantilever beam with tip loading

As it has been mentioned, in the present work the displacement and the strain of the structure are used as the input for NN training. Therefore, a validation of the analysis results of the strain and the displacement of the beam are needed. Table 1 shows the displacement of the cantilever beam. It can be seen that the displacements obtained in present work showed very good accordance compared with those obtained by MSC-Nastran and exact calculation. A very good accordance also shown by the strain of the beam obtained in the present work, by MSC-Nastran and exact calculation as depicted in Table 2. From comparisons in Table 1 and Table 2 it can be concluded that the results using FEM in present work are acceptable.

Table 1. Displacement of Cantilever Beam Used in Present Work

| Node | Present Result (m) | MSC-Nastran (m) | Exact (m) |
|---|---|---|---|
| 1 | 0 | 0 | 0 |
| 2 | $0.179 \times 10^{-4}$ | $0.182 \times 10^{-4}$ | $0.179 \times 10^{-4}$ |
| 3 | $0.688 \times 10^{-4}$ | $0.691 \times 10^{-4}$ | $0.688 \times 10^{-4}$ |
| 4 | $1.477 \times 10^{-4}$ | $1.483 \times 10^{-4}$ | $1.476 \times 10^{-4}$ |
| 5 | $2.500 \times 10^{-4}$ | $2.508 \times 10^{-4}$ | $2.500 \times 10^{-4}$ |
| 6 | $3.711 \times 10^{-4}$ | $3.720 \times 10^{-4}$ | $3.711 \times 10^{-4}$ |
| 7 | $5.063 \times 10^{-4}$ | $5.074 \times 10^{-4}$ | $5.062 \times 10^{-4}$ |
| 8 | $6.508 \times 10^{-4}$ | $6.521 \times 10^{-4}$ | $6.508 \times 10^{-4}$ |
| 9 | $8.000 \times 10^{-4}$ | $8.015 \times 10^{-4}$ | $8.000 \times 10^{-4}$ |

Table 2. Strain of cantilever beam used in present work

| Node | Present Result (m/m) | MSC-Nastran (m/m) | Exact (m/m) |
|---|---|---|---|
| 1 | $3.000 \times 10^{-4}$ | $2.999 \times 10^{-4}$ | $3.000 \times 10^{-4}$ |
| 2 | $2.625 \times 10^{-4}$ | $2.624 \times 10^{-4}$ | $2.625 \times 10^{-4}$ |
| 3 | $2.250 \times 10^{-4}$ | $2.249 \times 10^{-4}$ | $2.250 \times 10^{-4}$ |
| 4 | $1.875 \times 10^{-4}$ | $1.875 \times 10^{-4}$ | $1.875 \times 10^{-4}$ |
| 5 | $1.500 \times 10^{-4}$ | $1.499 \times 10^{-4}$ | $1.500 \times 10^{-4}$ |
| 6 | $1.125 \times 10^{-4}$ | $1.124 \times 10^{-4}$ | $1.125 \times 10^{-4}$ |
| 7 | $0.750 \times 10^{-4}$ | $0.749 \times 10^{-4}$ | $0.750 \times 10^{-4}$ |
| 8 | $0.375 \times 10^{-4}$ | $0.378 \times 10^{-4}$ | $0.375 \times 10^{-4}$ |
| 9 | 0 | 0 | 0 |

For training of NN the stiffness reduction factor *ee* was assumed to be random number from 0 to 1. The location of the damage in the structure is also assumed to be random. When the displacement as NN input, the training was conducted using different values of layer and β. Table 3 shows the training results of NN with displacement as input. Meanwhile, the variation of training error with number of epoche for several variations is shown in Fig. 5. It is interesting to note that the error for nndVar1 and nndVar2 get reduced significantly after a few epoch (300 epoche). However, it is observed that the variation of nndVar2 showed the best performance, the testing error become constant at $2.549 \times 10^{-5}$ after 600 epoche although the desired error ($1.000 \times 10^{-5}$) was not reached and it toke enough computational time .

Table 3. Variation of NN training and the result for displacement as input

| Variation | Code | Layer | β | MSE | Epoch | $t$ (s) |
|---|---|---|---|---|---|---|
| 1 | nndVar1 | 8 16 8 | 1 | $5.209 \times 10^{-5}$ | 1000 | 1679.3 |
| 2 | nndVar2 | 8 16 8 | 2 | $2.549 \times 10^{-5}$ | 1000 | 1653.6 |
| 3 | nndVar3 | 16 8 | 1 | $76.30 \times 10^{-5}$ | 1000 | 1289.3 |
| 4 | nndVar4 | 16 8 | 2 | $21.30 \times 10^{-5}$ | 1000 | 1259.5 |

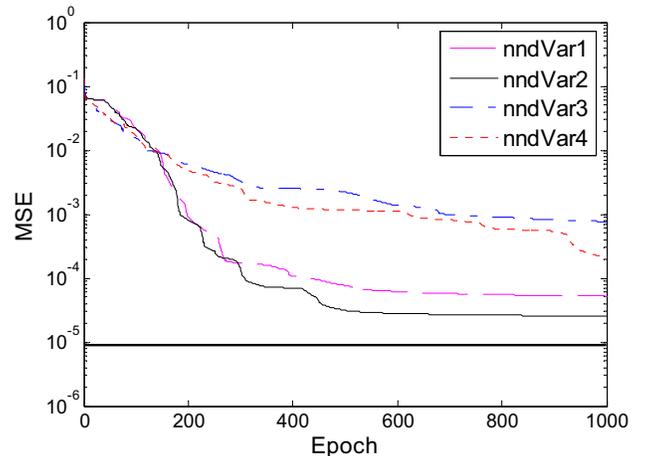

Fig. 5. Variation of training error with displacement as input

Strains of the beam are calculated at the same nodal point to that of displacement. Thus in this case also NN inputs have nine nodes consisting of strain values. The stiffness reduction factor in this case also to be assumed as random number from 0 to 1 with randomized location of damage. A variation as shown in Table 4 was applied to train the NN with strains as input. The result show that the desired error level ($1.000 \times 10^{-6}$) was reached by nnsVar2, nnsVar3 and nnsVar4 in range until 1000 epoche. However, it is observed from Fig.5 that the variation of nnsVar4 showed the best performance. The desired error may reached by this variation in 61 epoche with short time consuming.

Table 4. Variation of NN training and the result for strain as input

| Variation | Code | Layer | β | MSE | Epoch | $t$ (s) |
|---|---|---|---|---|---|---|
| 1 | nnsVar1 | 8 16 8 | 1 | $4.641 \times 10^{-5}$ | 1000 | 1665.7 |
| 2 | nnsVar2 | 8 16 8 | 2 | $0.981 \times 10^{-5}$ | 592 | 1002.7 |
| 3 | nnsVar3 | 16 8 | 1 | $0.987 \times 10^{-5}$ | 69 | 102.7 |
| 4 | nnsVar4 | 16 8 | 2 | $0.976 \times 10^{-5}$ | 61 | 83.4 |

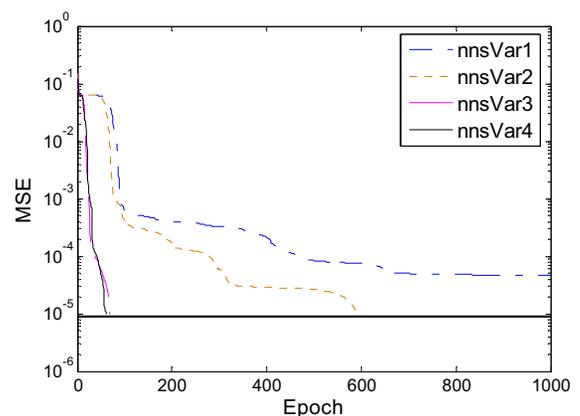

Fig. 6. Variation of training error with strain as input

From all the results can be seen that NN with strains as input can reach the desired error in small epoche and low computational time. More clearly, comparison results between input of strains and of displacement are depicted in Figs. 7, 8 and 9. Therefore, it will valueabel if the strains are applied as input in the NN training for the damage detection of structures or machinery components.

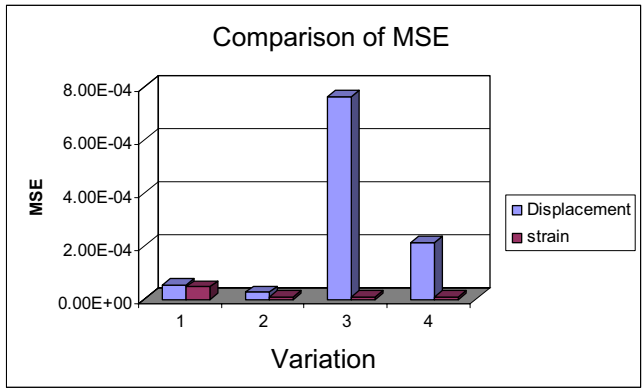

**Fig 7. Comparison of MSE for variation of NN training**

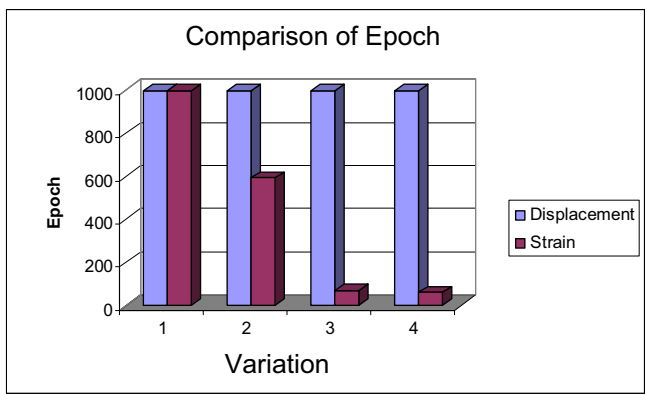

**Fig 8. Comparison of epoch for variation of NN training**

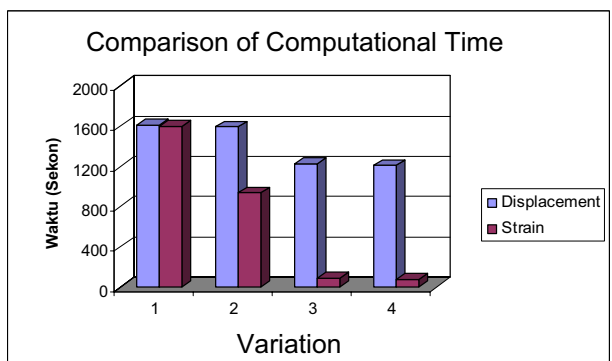

**Fig 9. Comparison of computational time for variation of NN training**

## 4  Conclusions

The main objective of the present research is to identify the size and the location of damage of a cantilever beam structure using neural network. For this purpose, a computer code was developed in which structural response in forms of displacement and strain due to damage is carried out. The damage of the structure is modeled as a stiffness reduction of a structural element wherein the stiffness reduction factor is treated as random number frm 0 and 1. In this research the location of the damage is also tretated as randomized location. The response data then are used as the input of the network to determine the size and the location of the damage. It is clearly observed from the result that selection of NN architecture is very important in the accuracy of the result. The networks showed that less hidden layer it gives shorter computational time. The factor of $\beta$ in training algorithm of Lavenberg Merquadrt will accelerate computational time and reaching of the desired error. The output results also showed that the performance of network improve when strain is used as input instead of displacement.

### References

[1] Bishop. C, "*Neural Networks for Pattern Recognition*," Oxford University Press, Oxford,1997.

[2] Demuth, Howard and Mark Beale,*" Neural Network Toolbox For Use with MATLAB*," Mathwork.Inc, USA, 1998.

[3] El-Sebakhy, E.A, "Artificial Neural Networks, Probabilistic Networks, Support Vector Machines, Adaptive-Neuro Fuzzy Systems, and Functional Networks," Elsevier Science, Saudi Arabia,2006.

[4] Hagan,M.T, and M. Menhaj, "*Training Feed forward Networks With The Marquardt Algorithm*," IEEE Transactions on Neural Networks, vol. 5, no. 6, pp. 989-993, 1994.

[5] Kusumadewi, Sri,"Membangun Jaringan Saraf Tiruan Menggunakan Matlab dan Excel Link", Graha Ilmu, Yogyakarta, 2004.

[6] Kwon, Y.W and Bang, H," *The Finite Element Method Using MATLAB*," CRC Press, Florida, 2000.

[7] Maity, D and Saha, A,"Damage Assessment in Structure from Changes in Static Parameter Using Neural Network," Sadhana Vol 29, Part 3, pp.315-327, June 2004.

[8] Siang, J.J, "*Jaringan Syaraf Tiruan*," Penerbit Andi, Yogyakarta, 2005.

[9] Yijun, L,"Lectures Notes: Introduction to Finite Element Method," University of Cincinnati, USA, 1998.